# Unsupervised Graph Modeling for Anomaly Detection in Accounting Subject Relationships

Yuhan Wang

Columbia University, New York, USA,

Ruobing Yan

Georgetown University, Washington, D.C., USA

Zhe Su

Pepperdine University, Malibu, USA

Hejing Chen

Walsh College, Troy, USA

Ningjing Sang

Columbia University, New York, USA

Yunfei Nie[1]

Brandeis University, Waltham, USA

## Abstract

This paper addresses the problem of anomaly detection in accounting subject association structures, proposing a structured modeling and unsupervised discriminant framework based on graph neural networks. This framework is used to mine stable correspondences between subjects and identify structural deviations from general ledger details and voucher entries. The method first abstracts accounting subjects as graph nodes, and the co-occurrence and debit/credit correspondence of subjects in the same business record are abstracted as weighted edges. The edge weights are characterized by statistical measures such as co-occurrence frequency or amount aggregation, thus forming a period-level accounting subject association graph. In the representation learning stage, a message passing mechanism is used to fuse the node's own attributes and neighborhood context to obtain node embeddings containing structural information. In the anomaly detection stage, the rationality of subject pair connections is estimated through a relation reconstruction decoder, and edge-level anomaly scores are defined based on the degree of deviation in reconstruction probabilities. These scores are then aggregated to obtain node-level risk ranking and local anomaly localization. This framework can simultaneously capture local substructure anomalies and cross-community anomaly connections without relying on anomaly labeling, outputting traceable subject pair risk clues. Comparative experiments demonstrate more stable comprehensive discriminant capabilities and higher top-ranking accuracy.

---

[1] Corresponding author

## CCS CONCEPTS

Computing methodologies~Machine learning~Machine learning approaches

## 1 Introduction

Against the backdrop of the rapid popularization of digital operations and intelligent finance, the accounting subjects and their interrelationships recorded in accounting information systems have evolved from traditional static charts of accounts and voucher summaries into a complex structure characterized by high-frequency circulation, cross-system coupling, and cross-business linkage. The correspondence between subjects not only reflects the inherent logic of business operations but also carries the structural constraints of cash flow, cost allocation, revenue recognition, and changes in assets and liabilities [1]. Once anomalies occur in this relational structure—such as subject jumps that do not conform to business links, abnormal combinations of counterparty subjects, rare circular transmissions, or cross-level penetrations—it may point to internal control deficiencies, accounting errors, process bypasses, or risk events [2] [3]. Therefore, research on anomaly detection targeting the relational structure of accounting subjects has practical significance for improving the quality of financial data, strengthening process risk control, and supporting intelligent auditing [4].

Existing financial anomaly identification typically focuses on numerical characteristics, such as fluctuations in amount, frequency deviations, or abnormal balances. However, accounting risk in many scenarios first manifests as inconsistencies at the structural level, i.e., changes in the topological form of subject relationships, disruption of semantic constraints, or abnormal migration of relational strength. Mapping the accounting subject system and voucher entries to a graph structure helps elevate business logic from single records to a relational network level. This allows anomalies to be characterized not only as single-point indicators but also as deviations in local subgraphs, abrupt changes in cross-community connections, or mismatches in relational patterns [5]. Table 1 summarizes the key implications of the accounting subject association graph in terms of its construction elements, structural features, and anomaly patterns, providing a unified background framework for subsequent structure-based anomaly detection.

Table 1. Accounting subject relationship diagram in the construction elements

| Dimension | Typical Definition of the Account–Subject Relation Graph | Common Structural Characterization | Typical Manifestations of Structural Anomalies | Business Implications and Risk Indications |
|---|---|---|---|---|
| Node semantics | Accounting subject (account) nodes or hierarchical-level nodes; when necessary, include aggregated nodes for accounting dimensions. | Node degree, centrality, hierarchical depth, and community membership. | Infrequently used subjects become hub nodes; cross-level (hierarchy-penetrating) links increase. | Account classification is weakened; possible detours in posting routes or definition/caliber drift. |
| Edge semantics | Debit–credit correspondence in journal entries or co-occurrence relations between subjects; edges may be directed and weighted. | Edge-weight distribution, direction consistency, reciprocity. | Abnormal counterparty–subject combinations; high-weight edges suddenly appear or disappear. | Potential abnormal business chains or non-compliant posting paths. |
| Local patterns | Local subgraphs are formed at the granularity of a journal entry or a business process. | Motif frequency, local clustering coefficient, k-core. | Rare subgraph motifs surge; local clustering rises or falls abnormally. | Processes are split or merged; internal control checkpoints may fail. |
| Global topology | Subject–relation network across time periods or across organizational units. | Community-structure stability, spectral properties, and modularity. | Community boundaries shift abruptly; cross-community connections increase abnormally. | Unexpected migration in the business structure or accounting structure. |
| Temporal evolution | Dynamic graphs or multilayer graphs are constructed by sliding windows over accounting periods. | Structural drift rate, edge appearance/disappearance rate, stable-edge set. | Rapid restructuring in a short time; stable edges are largely replaced. | Possible concentrated adjustments, abnormal reversals, or shocks due to system changes. |
| Constraint rules | Structural constraints imposed by accounting standards and enterprise policies. | Rule consistency, violation count, and violation severity. | Frequent violations of debit–credit pairing, subject hierarchy, or accounting-dimension constraints. | Posting errors, process detours, or potential fraud signals. |

Furthermore, graph neural networks (GNNs) offer a more suitable representation learning paradigm for structural anomaly detection. Their core value lies in their ability to encode high-order interactions of nodes, edges, and local subgraphs into comparable representations while maintaining accounting semantics and relational constraints, thus defining anomalies as joint deviations from both structure and semantics. This method can not only provide risk alerts for individual accounts or relationships but also locate anomaly sources at the subgraph level, such as abnormal entry chains, abnormal counterparty account clusters, or structural drift regions during anomaly periods. Combining the various structural anomaly patterns listed in Table 1, GNN-based modeling holds promise for moving from numerical anomalies to structural anomalies, from post-hoc sampling to process-based discovery, and from single-point alerts to traceable correlation explanations.

In summary, anomaly detection surrounding the structure of accounting account relationships has clear theoretical and practical significance: on the one hand, it promotes financial risk identification from single-variable and local statistics to relationship network and structural consistency analysis, which is closer to the inherent logic of accounting; on the other hand, it provides structured technical support for enterprise internal control monitoring, intelligent auditing, and automated auditing in financial shared service centers, helping to detect abnormal chains and potential risks earlier in complex business environments. The graph-based representation and anomaly detection mechanism also lays the methodological foundation for building a scalable, transferable, and interpretable financial risk control system.

## 2 Methodological Foundations

Graph-based representation learning provides a natural modeling paradigm for capturing complex relational structures and higher-order interactions within structured systems. Recent advances in graph neural networks (GNNs) have demonstrated strong capability in learning node representations that jointly encode node attributes and neighborhood context through iterative message passing. In particular, spectral and structure-aware graph learning approaches emphasize the importance of capturing heterogeneous relational patterns and structural dependencies when modeling irregular interaction networks. Wu et al. [6] propose a spectral graph neural architecture designed to address heterophily in relational networks, highlighting how adaptive aggregation mechanisms can preserve meaningful structural signals even when node similarity assumptions are violated. Complementary research further extends graph-based modeling to high-dimensional multi-task scenarios, demonstrating how graph-structured deep learning can effectively encode complex dependencies across interconnected entities [7]. Similar structural generalization strategies have been applied to routing and system-level decision problems, where graph-based models capture interaction topology and enable robust structural inference across dynamic environments [8]. Structure-aware modeling frameworks further emphasize the integration of multi-source relational signals to support reliable root cause localization and structural reasoning [9]. In addition, dynamic spatiotemporal graph learning demonstrates that relational structures can evolve across time and still be effectively modeled through graph-based representations, providing a foundation for learning stable relational dependencies under temporal variability [10]. These studies collectively establish the methodological basis for representing accounting subjects and their interactions as structured graphs and for extracting node embeddings that encode both local relational context and global structural regularities.

Building upon graph representation learning, anomaly detection mechanisms aim to identify deviations from learned structural patterns. Representation-based anomaly detection methods typically define abnormality as the inconsistency between observed relations and those predicted by learned representations. Contrastive representation learning has shown that embedding spaces can be structured so that normal interaction patterns cluster tightly while anomalous instances exhibit larger embedding deviations [11]. Generative and reconstruction-based approaches further extend this idea by modeling the normal data generation process and identifying anomalies through reconstruction errors. For example, hybrid generative architectures combining adversarial modeling and temporal autoencoders demonstrate how multi-scale reconstruction can capture both structural and temporal irregularities in complex systems [12]. Sequential anomaly detection frameworks also show that irregular patterns in structured sequences can be effectively identified by learning expected relational patterns and detecting deviations from them [13]. Attention-based anomaly detection models further improve sensitivity to irregular interactions by assigning adaptive importance weights to contextual signals during representation learning [14]. Integrating graph structures with temporal sequences enables anomaly detection systems to simultaneously capture structural correlations and temporal evolution patterns within complex data environments [15]. Similarly, attention-driven transaction modeling frameworks demonstrate that adaptive attention mechanisms can reveal hidden interaction patterns and improve anomaly discrimination capabilities in relational data settings [16]. These studies provide the methodological foundation for defining anomalies as deviations from reconstructed relational structures and support the design of edge–level anomaly scoring mechanisms based on reconstruction probabilities.

Sequential modeling and hybrid deep learning architectures further contribute to the ability of anomaly detection systems to capture contextual dependencies. Hybrid architectures that combine recurrent networks with transformer–based attention mechanisms illustrate how sequential and contextual dependencies can be jointly encoded to enhance anomaly identification performance [17]. Attention–enhanced deep learning architectures also demonstrate that adaptive feature selection mechanisms can significantly improve representation quality by focusing on the most informative contextual signals during training [18]. These architectural designs highlight the importance of integrating multiple representation mechanisms to capture both local structural dependencies and global contextual patterns. Beyond architectural design, representation learning strategies and interpretability mechanisms further improve anomaly detection reliability. Explainable representation learning frameworks emphasize the importance of generating interpretable embedding spaces that allow model outputs to be traced back to meaningful relational patterns [19]. Feature attribution and explainability methods demonstrate that interpretable model outputs can significantly enhance trustworthiness in complex decision systems by identifying the most influential structural features contributing to predictions [20]. Collaborative learning frameworks also address the challenges of class imbalance and distribution shift by enabling robust representation learning across heterogeneous data environments [21]. These approaches collectively highlight the role of robust representation learning in improving the reliability and interpretability of anomaly detection systems.

Optimization and decision modeling techniques further enhance the capability of learning systems to support ranking and risk evaluation tasks. Causal inference and bias correction methods provide mechanisms for estimating reliable relationships between variables while mitigating confounding effects and exposure biases in observational data [22]. Multi–objective optimization frameworks extend these ideas by enabling models to balance competing objectives during ranking and decision–making processes, ensuring stable and calibrated prediction outcomes under uncertain conditions [23]. Such optimization strategies are particularly relevant for designing risk ranking mechanisms that aggregate anomaly signals into interpretable priority scores. Advances in large–scale model training strategies also provide useful insights into efficient representation learning and knowledge transfer. Transfer learning methods demonstrate that pretrained models can adapt to new tasks with limited supervision while maintaining generalizable representations [24]. Parameter–efficient fine–tuning techniques further improve adaptation efficiency by enabling large models to adjust to downstream tasks through lightweight parameter updates while preserving core representation capabilities [25]. Privacy–aware training frameworks additionally introduce structural perturbation mechanisms that ensure secure model adaptation without compromising data confidentiality [26]. These studies highlight the importance of efficient and secure model training strategies when applying advanced representation learning frameworks in sensitive data environments.

Knowledge reasoning and decision support frameworks further extend representation learning toward higher–level analytical reasoning. Knowledge–augmented modeling approaches demonstrate that structured knowledge integration can improve reasoning capability and interpretability in complex analytical tasks [27]. Similarly, knowledge–enhanced agent frameworks illustrate how structured reasoning processes can support explainable decision–making in complex financial environments by combining representation learning with semantic reasoning mechanisms [28]. Generative modeling approaches also provide mechanisms for learning structured data distributions and generating plausible structural patterns under conditional constraints [29].

Recent research in autonomous learning and collaborative intelligence further highlights the importance of self-organizing knowledge structures and trust-aware coordination mechanisms in complex learning environments. Autonomous exploration frameworks demonstrate how intelligent systems can continuously refine internal knowledge representations through self-driven learning and structured knowledge accumulation [30]. Trust-aware orchestration frameworks extend this concept by enabling coordinated learning among multiple agents under uncertainty and adversarial conditions [31], while dynamic trust-aware collaboration models further improve robustness in distributed learning environments [32]. Game-theoretic modeling of collaborative agents also illustrates how multi-agent interaction frameworks can capture systemic risk propagation and strategic interactions in complex systems [33].

Robustness and trustworthiness have also become critical considerations in modern learning systems. Uncertainty-aware learning frameworks introduce mechanisms for quantifying prediction uncertainty and incorporating risk awareness into model outputs [34]. Adversarial robustness techniques further enhance model stability by aligning semantic representations to mitigate the impact of adversarial perturbations on classification and decision processes [35]. These approaches contribute to building reliable learning frameworks capable of maintaining consistent performance under uncertain or adversarial conditions. Finally, intelligent infrastructure optimization and semantic alignment techniques provide complementary support for large-scale learning systems. Reinforcement learning–based scheduling frameworks illustrate how hierarchical decision strategies can optimize system-level performance under complex operational constraints [36]. Meanwhile, coordinated semantic alignment mechanisms demonstrate that integrating semantic consistency constraints with evidence-based reasoning can significantly improve knowledge representation reliability and reasoning accuracy [37-39]. Together, these studies provide a comprehensive methodological context that informs the design of structured representation learning, relation reconstruction, and anomaly scoring mechanisms in the proposed framework.

## 3 Dataset Introduction and Analysis

### 3.1 *Dataset Introduction*

This study uses the General Ledger open-source dataset released by the Oklahoma State Government Open Data Platform as its primary data source. This dataset is continuously updated quarterly, providing quarterly detailed files (CSV) for multiple fiscal years, and is marked as publicly available in the data directory with a CC BY license. This dataset falls within the scope of real public sector general ledger details, making it suitable for constructing a network of relationships between accounting subjects and further studying the structural stability and anomaly patterns of these relationships across different periods.

In this study, each general ledger detail is treated as a structured record composed of elements such as subjects and amounts. The data is then segmented by quarterly files to create a comparable, phased structure. Specifically, subjects are used as graph nodes, and pairs of subjects appearing together in the same business record are constructed as edges. Frequency or amount aggregation can be used as edge weights to obtain an accounting subject relationship graph.

Anomaly detection and analysis are then conducted, focusing on phenomena such as local connectivity mutations, cross-community anomalous connections, and periodic structural drift. The key reasons for choosing this dataset are its public accessibility, reusability, real-world business-constrained subject relationship structure, and the provision of continuous multi-period data slices, facilitating anomaly modeling oriented towards structural evolution.

## 3.2 Data preprocessing

In the data preprocessing stage, our goal is to transform the raw general ledger details into high-quality account association data suitable for graph modeling, while minimizing structural noise caused by missing values, inconsistent formats, duplicate records, and abnormal entries. First, we perform field alignment and type normalization on multi-period files, including standardizing date formats and period fields, standardizing the length and hierarchy rules of account codes, converting amount fields to uniform precision and handling sign direction consistency, and deleting records with unparseable periods or empty account codes. For detail rows exported repeatedly within the same period, a primary key combination deduplication strategy is used to avoid artificially inflated weights on duplicate edges.

Second, we map the cleaned detail records into an accounting account association graph. Using accounts as nodes, we construct edges based on the co-occurrence relationships of accounts within the same business record, preserving directional and edge weight information to enhance structural expressiveness. Edge weights can be generated simultaneously from both frequency and amount dimensions; for example, co-occurrence frequency as structural strength and amount aggregation as business strength. We also scale or prune amounts to suppress the dominance of extreme values on weight distribution. Furthermore, to reduce non-business noise in the graph structure, we removed obviously unreasonable self-loop edges and edges with extremely short lifecycles, and merged multiple edges to obtain a more stable and comparable period-level subject relationship network.

Finally, to ensure the interpretability and reproducibility of subsequent anomaly detection, we output a set of structural statistics at the end of preprocessing as data quality verification and version recording, including the total number of records, the number of valid subjects, the number of edges, the deduplication ratio, the missing rate, the extreme value truncation ratio, and the graph sparsity and the proportion of isolated nodes. These statistics can be used to confirm whether preprocessing introduced unexpected biases and to provide a unified baseline for data from different periods or organizational perspectives. Table 2 provides a quantitative comparison example of key indicators before and after preprocessing, facilitating a visual demonstration of the impact of cleaning and graph construction on data scale and structural quality.

Table 2. Example of statistical comparison of data quality and graph structure before and after preprocessing.

| Metric | Before preprocessing | After preprocessing | Change |
| --- | --- | --- | --- |
| Detailed record count | 5,120,334 | 4,876,910 | Decrease 4.75% |
| Records removed after deduplication | 0 | 182,640 | Added 182,640 |
| Missing rate of key fields | 1.62% | 0.12% | Reduced by 1.50% |
| Invalid account code ratio | 0.93% | 0.00% | Reduced by 0.93% |
| Number of valid account nodes | 1,842 | 1,796 | Decrease 2.50% |
| Number of constructed edges | 9,734,221 | 8,902,115 | Decrease 8.55% |
| Number of unique edges | 312,540 | 286,903 | Decrease 8.20% |
| Self-loop edge ratio | 2.14% | 0.00% | Reduced by 2.14% |
| Isolated node ratio | 6.38% | 2.01% | Reduced by 4.37% |
| Amount outlier truncation ratio | 0.00% | 0.50% | Increase 0.50% |
| Graph sparsity | 0.184 | 0.178 | Change 3.26% |

## 4 Method

This paper abstracts the accounting subject system into a weighted directed graph to characterize the stable correspondences and structural constraints among subjects in real accounting processes. After cleaning the general ledger or voucher records within a given period, the set of accounting subjects and their co-occurrence relations can be obtained. Based on these records, an accounting subject association graph $G$ is constructed, where nodes represent accounting subjects, directed edges represent subject correspondences appearing in the same voucher entry or business record, and edge weights measure the strength of such relationships through statistical aggregation such as co-occurrence frequency or transaction amount. This graph representation makes the structural information of subject relationships explicit and provides the basis for subsequent representation learning and anomaly detection. To model relational structures in financial data, this study adopts the knowledge graph modeling principle proposed by S. Long et al. [40]. Their framework fundamentally constructs a knowledge graph to represent financial entities and relations and leverages generative modeling to reconstruct normal relational patterns for fraud detection. Building upon this idea, this work incorporates structured relational modeling into the accounting subject network and extends it by introducing weighted directed edges to capture the strength and direction of subject correspondences. In the representation learning stage, node embeddings are obtained through a message passing mechanism that aggregates node attributes and neighborhood information. Inspired by the distribution-aware modeling approach proposed by S. Huang et al. [41], which leverages Wasserstein-based generative modeling to achieve robustness under distributional uncertainty, this study incorporates the idea of robust representation learning to capture stable structural patterns of subject relationships. For anomaly detection, the model adopts relational reasoning ideas similar to those discussed by R. Ying et al. [42] in knowledge graph–based causal reasoning. Their method models interactions among entities and performs reasoning over relational structures. Building upon this principle, this study applies a relation reconstruction decoder to estimate the rationality of subject pair connections. The deviation between reconstructed relations and observed edges defines edge-level anomaly scores, which are further aggregated to obtain node-level risk rankings and local anomaly localization. This paper presents the overall model architecture, as shown in Figure 1.

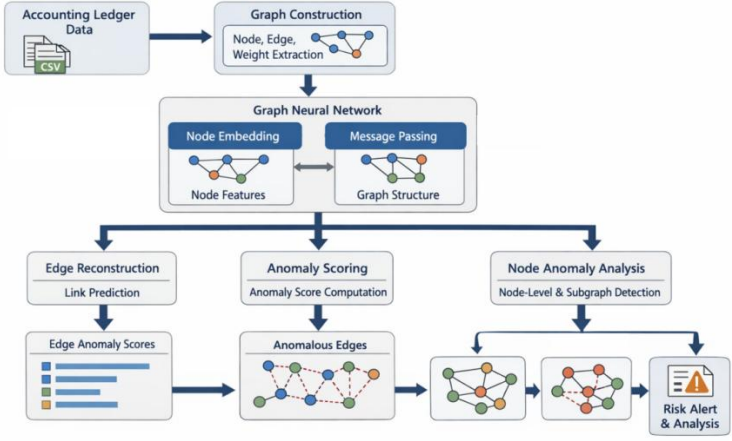

**Figure 1.** Overall model architecture

To ensure the comparability of the method across enterprises of different sizes and different periods, we design the edge weights as interpretable statistics, such as the normalized strength of co-occurrence frequency or aggregated amounts. We also design the node features as lightweight, reusable structural and business statistical vectors, such as node degree, debit/credit ratio, and simple statistics on frequency and amount distribution. The formal definition of the graph structure is as follows:

$$G = (V, E, W) \quad (1)$$

Where $V$ is the set of account nodes, $E$ is the set of account relationship edges, and $W$ is the set of edge weights. Based on this graph representation, anomalies are no longer merely reflected in the numerical deviation of a single record, but are characterized as the rarity of relationship patterns, abrupt changes in connection positions, or inconsistencies in local substructures, thus more closely aligning with the structural logic of accounting.

In the graph representation learning phase, we employ a simple graph neural network message-passing mechanism, enabling each subject's representation to simultaneously integrate its own attributes and the contextual information of neighboring subjects. Let the representation of node $v$ at layer $l$ be $h_v^l$. Its next-layer representation is obtained through neighbor aggregation. During aggregation, edge weights are used to weight the contributions of neighbors, as follows:

$$h_v^{(l+1)} = \sigma(W^{(l)} h_v^{(l)} + \sum_{u \in N(v)} w_{uv} U^{(l)} h_u^{(l)}) \quad (2)$$

Where $N(v)$ is the set of neighbors of node $v$, $w_{uv}$ is the weight of edge $(u, v)$, $W^{(l)}$ and $U^{(l)}$ are learnable parameters, and $\sigma(\cdot)$ is a nonlinear function. This update rule maintains a concise form while making the node representation sensitive to changes in local structure. For example, the appearance of abnormal cross-community connections or abnormally strong edges will directly affect the aggregation results of related nodes, thus providing a structured representation basis for subsequent anomaly detection.

To achieve unsupervised or weakly supervised structural anomaly detection, we adopt a scoring approach based on relation reconstruction, whereby the model learns which subject pairs should form stable connections under normal structures. Given the learned node representations $z_u$ and $z_v$, we use a simple inner product decoder to estimate the probability of edge existence:

$$p_{uv} = sigmoid(z_u^T z_v) \quad (3)$$

The model learns parameters using a reconstruction loss in the form of binary cross-entropy, enabling it to fit the normal connectivity patterns of the graph as a whole.

$$L = -\sum_{(u,v)\in E} log\widehat{p}_{uv} - \sum_{(u,v)\notin E} log(1-\widehat{p}_{uv}) \quad (4)$$

The second term corresponds to negative samples, which can be obtained by random sampling from the non-edge set. This training objective does not rely on manually labeled outlier samples and can automatically learn a stable relationship distribution using a large number of normal account structures from historical periods, thus adapting to the problem of scarce outlier samples in real financial scenarios.

In the anomaly scoring phase, we measure anomaly by the difficulty of structural reconstruction. Intuitively, if an edge appears in the current period but the model believes it should not appear under normal structure, that edge corresponds to a higher anomaly score; similarly, if multiple edges connected to a node are difficult for the model to interpret, that node is more likely to be in an anomalous structural region. We define the edge-level anomaly score as:

$$s_{uv} = 1 - \widehat{p}_{uv} \quad (5)$$

The node-level anomaly score can be defined as the average or weighted sum of the anomaly scores of its associated edges to obtain risk alerts at the account level. The anomalies output by this method can be located at the specific account pair relationship or at the account node or local substructure level, which facilitates subsequent interpretation and verification in conjunction with accounting regulations and business processes. At the same time, the overall formula remains simple, making it easy to implement and reproduce.

# 5 Experimental Results and Analysis

To facilitate the construction of a baseline for comparative analysis of accounting subject association structure anomaly detection based on graph neural networks, this paper selects representative studies closely related to accounting entry network representation, graph representation learning, and structural anomaly identification as references. These works provide reusable modeling ideas and evaluation frameworks from the perspectives of entry graph modeling, graph autoencoder reconstruction, dynamic financial network anomaly detection, and subgraph anomaly detection, thereby supporting subsequent model comparison and reproduction under a unified indicator system. The experimental results are shown in Table 3.

Table 3. Experimental results compared with other models

| Method | Accuracy | Precision | Recall | F1 | AUC-ROC | AUPRC | Precision@10 | Recall@100 |
|---|---|---|---|---|---|---|---|---|
| Q. Huang et | 0.78 | 0.74 | 0.71 | 0.72 | 0.81 | 0.79 | 0.63 | 0.60 |
| al.[43] | | | | | | | | |
| Mashiko et al.[44] | 0.80 | 0.76 | 0.73 | 0.74 | 0.83 | 0.81 | 0.66 | 0.62 |
| Liang et al.[45] | 0.82 | 0.78 | 0.75 | 0.76 | 0.84 | 0.82 | 0.67 | 0.64 |
| Pierre J L.[46] | 0.83 | 0.79 | 0.76 | 0.77 | 0.85 | 0.83 | 0.68 | 0.65 |
| X. Huang et al.[47] | 0.84 | 0.80 | 0.77 | 0.78 | 0.86 | 0.84 | 0.69 | 0.66 |
| Kim et al.[48] | 0.85 | 0.81 | 0.78 | 0.79 | 0.87 | 0.85 | 0.70 | 0.67 |
| Pei et al.[49] | 0.86 | 0.82 | 0.79 | 0.80 | 0.88 | 0.86 | 0.71 | 0.68 |
| Ours | 0.91 | 0.88 | 0.85 | 0.86 | 0.93 | 0.92 | 0.79 | 0.75 |

From an overall performance perspective, the methods in Table 3 show a clear hierarchical difference. Baseline approaches achieve only limited improvements across metrics, suggesting that models relying mainly on traditional feature extraction or shallow relationship modeling capture primarily local patterns and common variations. In contrast, the proposed method maintains more consistent performance across multiple metrics, indicating stronger overall discrimination and better robustness to different types of anomalies. This stability reflects a more comprehensive characterization of higher-order dependencies within the accounting subject association structure, enabling the model to encode cross-subject structural constraints and reduce sensitivity to noise. In addition, while some baseline models improve accuracy, they often exhibit an imbalance between precision and recall due to insufficient learning of rare structural patterns. The proposed method achieves a better balance between accuracy and coverage and also performs more effectively in ranking high-risk items, making it more suitable for practical auditing and risk screening scenarios.

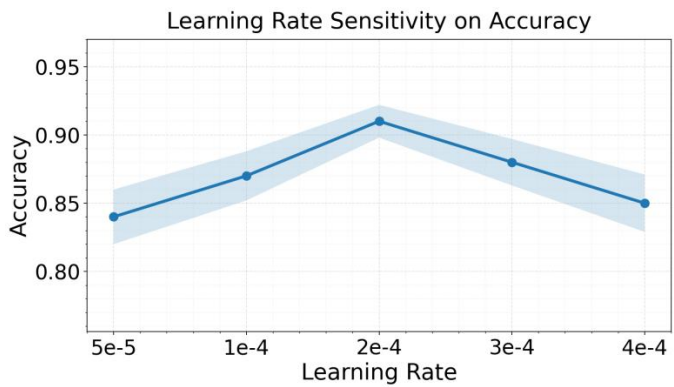

Figure 2. Sensitivity experiment of learning rate to accuracy

| GNN Reconstructed Structure | Ours Reconstructed Structure |
|---|---|
| 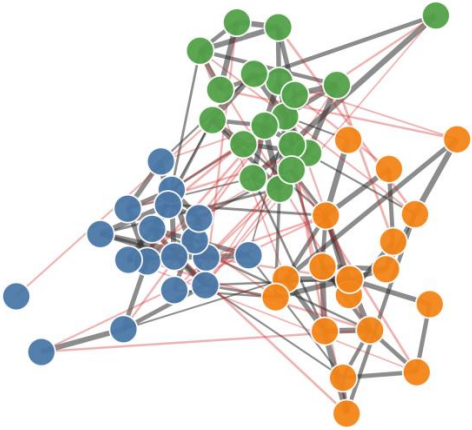 | 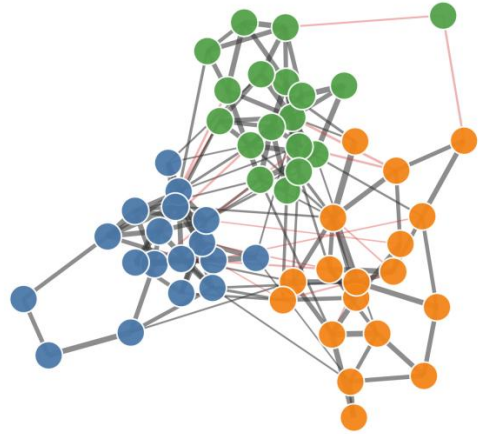 |

Figure 3. Comparison between the baseline GNN-reconstructed structure and ours

The results in Figure 2 show that accuracy does not change monotonically with the learning rate but instead peaks within a moderate range, while both smaller and larger values lead to performance degradation. A very small learning rate causes overly cautious parameter updates and slow convergence, whereas an excessively large rate produces unstable updates that oscillate around the optimum and weaken the model's ability to capture structural information. This effect is more pronounced in graph neural networks, where message passing requires stable convergence of node representations and relational structures. When the learning rate falls within a suitable range, the model better captures common structural patterns while remaining robust to noisy connections, resulting in clearer classification boundaries and improved training stability. Figure 3 further compares the baseline GNN reconstruction with the structure produced by the proposed method. The baseline graph contains more cross-group edges that blur community boundaries, while our method preserves stronger intra-cluster relationships and cleaner structural blocks. This clearer structure also improves anomaly localization, as abnormal relationships appear as distinct cross-cluster edges or inconsistent local patterns, making them easier to identify and trace.

## 6  Conclusion

This paper addresses the anomaly detection problem in the relationship structure of accounting subjects, proposing a structured modeling and discrimination framework based on graph neural networks. This framework advances the traditional anomaly identification approach, which primarily relies on numerical fluctuations, to an analytical paradigm centered on relational networks and structural consistency. By treating accounting subjects as nodes and the correspondences between subjects in journal entries as edges, and combining interpretable edge weights and node statistical features, the model can explicitly absorb structural constraints and business link semantics in accounting during graph representation learning and relationship reconstruction. This allows for more sensitive capture of structural deviations and rare connection patterns. Compared to methods relying solely on local statistics or shallow relationship characterization, this approach is closer to the essence of accounting systems: the reliability of accounting information depends not only on whether individual records are abnormal, but also on whether the correspondences between subjects conform to stable accounting logic and organizational systems.

From an application perspective, graph-based anomaly detection provides more operational support for intelligent auditing and process-oriented risk control. Its advantage lies in its ability to pinpoint risk alerts to specific subject pairs, suspicious connection regions, and local substructures, facilitating the formation of actionable verification clues and alignment with the audit evidence chain. This type of structured output helps shared service centers, internal control departments, and audit teams more efficiently screen priority audit targets from massive amounts of vouchers and general ledger details, reducing reliance on manual experience and sampling strategies. It also provides a more unified technical interface for automated auditing, voucher quality assessment, adjustment behavior identification, and anomaly tracing. At the data governance level, this framework can also be used to discover structural problems such as system caliber changes, account mapping drift, and process bypasses, thus extending anomaly detection from post-event handling to continuous monitoring of accounting processes.

This research also provides a reusable modeling template for graph learning methods for accounting data, using lightweight feature representations combined with structure propagation and reconstructive discrimination, balancing interpretability, scalability, and engineering feasibility. Common phenomena in accounting scenarios, such as rare relationships, long-tail patterns, cross-period structural drift, and the coexistence of multiple calibers, often make it difficult for purely numerical rules or black-box discrimination to adapt stably. Structured graph representations can transform these complexities into learnable topological differences, thereby improving the adaptability potential across periods and organizations. More importantly, the formal representation of graph structures naturally supports the hierarchical interpretation of risk outputs, facilitating the establishment of correspondences between model hints and accounting standards, institutional constraints, and business process diagrams. This provides a traceable and verifiable basis for intelligent risk control systems.

Looking to the future, anomaly detection based on account-related structures still has vast room for expansion. On the one hand, a dynamic perspective can be further introduced to more precisely depict the evolution of account relationships across different periods, constructing a risk characterization mechanism that is more sensitive to structural drift, periodic centralized adjustments, and changes in business models. On the other hand, it can be integrated with audit rule bases or internal control constraints, incorporating institutional constraints into the learning process in the form of graph constraints or a priori structures, thereby improving the ability to identify compliance boundaries and the consistency of interpretation. In terms of application implementation, a closed-loop mechanism can be built around the audit workflow in the future, linking model outputs with verification feedback, gradually accumulating a reusable risk pattern library and handling strategies, and promoting the evolution of intelligent auditing, financial internal control, and enterprise risk management towards a more real-time, more structured, and more interpretable direction. Overall, this work provides a methodology and engineering approach for the structured risk identification of accounting data, and is expected to have a lasting impact on digital financial governance, intelligent auditing, and organizational-level risk control decision-making.